# DeepIrisNet2: Learning Deep-IrisCodes from Scratch for Segmentation-Robust Visible Wavelength and Near Infrared Iris Recognition

Abhishek Gangwar, Akanksha Joshi, Padmaja Joshi
Centre for Development of advanced Computing
(CDAC), Mumbai, India
*{abhishek,akanksha,padmaja}@cdac.in*

R. Raghavendra
Norwegian Biometrics Laboratory, Norwegian
University of Science and Technology (NTNU),
2802 Gjøvik, Norway
*raghavendra.ramachandra@ntnu.no*

## Abstract

*We first, introduce a deep learning based framework named as DeepIrisNet2 for visible spectrum and NIR Iris representation. The framework can work without classical iris normalization step or very accurate iris segmentation; allowing to work under non-ideal situation. The framework contains spatial transformer layers to handle deformation and supervision branches after certain intermediate layers to mitigate overfitting. In addition, we present a dual CNN iris segmentation pipeline comprising of a iris/pupil bounding boxes detection network and a semantic pixel-wise segmentation network. Furthermore, to get compact templates, we present a strategy to generate binary iris codes using DeepIrisNet2. Since, no ground truth dataset are available for CNN training for iris segmentation, We build large scale hand labeled datasets and make them public; i) iris, pupil bounding boxes, ii) labeled iris texture. The networks are evaluated on challenging ND-IRIS-0405, UBIRIS.v2, MICHE-I, and CASIA v4 Interval datasets. Proposed approach significantly improves the state-of-the-art and achieve outstanding performance surpassing all previous methods.*

## 1. Introduction

In more recent years, iris recognition for non-ideal irises has gained substantial importance to facilitate iris recognition under less constrained environmental conditions [9], from a distance and with less user cooperation. Following the trend, various options have been explored such as iris recognition in Visible Spectrum (VS) and iris recognition on mobile phone devices etc.. Key advantages of visible spectrum iris recognition include the possibility of iris imaging in on-the-move and at-a-distance scenarios as compared to fixed range imaging in near-infra-red light. Well known challenges like Noisy Iris Challenge Evaluation (NICE I, II) Iris biometrics competition [2] and Mobile Iris CHallenge Evaluation" (MICHE I, II) [6] reiterate the increasing interest in visible spectrum iris recognition. Also, recent mobile phones have cameras at par with the imaging capabilities of dedicated digital cameras which makes them viable for iris recognition. Biometrics authentication over mobile phone devices will extend functionality and capabilities of traditional biometric identification systems.

The major focus of this paper is to present an iris recognition framework to break through the practicality limitations of the existing iris recognition systems. In particular, we propose a novel and robust deep CNN based iris recognition framework which can work effectively for unconstrained iris acquisition in NIR or visible spectrum. Due to unconstrained nature of image capture, the images will pose challenges in accurate iris segmentation and may not have good quality iris texture to be used for recognition. In order to answer these issues, the proposed framework introduces the following: i) an iris recognition framework which is highly tolerant to segmentation inaccuracies and we show that it is possible to do highly accurate iris recognition without performing traditional Daugman's Rubber Sheet model based iris normalization [3], ii) a new deep CNN based architecture referred to as DeepIrisNet2 for iris texture representation in an attempt to extract effective and compact iris features.

The major contributions in this paper can be summarized as follows-

- We first present a simple yet effective supervised learning framework named as DeepIrisNet2 to obtain highly effective iris representation such that irises can be recognized with invariance to illumination, scale, and affine transformations. The proposed CNN is trained from scratch and we provide extensive details about training hyper parameters, along with intuition for their selection to publish fully reproducible CNN for iris recognition.

- Unlike traditional iris recognition systems, proposed DeepIrisNet2 is trained directly using rectilinear iris image without normalization using Daugman's Rubber Sheet Model to convert Cartesian to polar coordinate system.

- We present a novel dual CNN based iris segmentation framework. The first stage CNN detects iris and pupil bounding boxes and second stage CNN takes the region detected by bounding boxes as input and performs a semantic pixel-wise segmentation to obtain actual iris region. The process decreases the computational cost and alleviates the impact of occlusions and other noise.

- We build a large scale hand labeled dataset named as

Iris Detection and Segmentation Groundtruth Database (IrisDSGD) comprising of; annotations for iris, pupil bounding boxes and pixel level labeling of iris region. The dataset is used to train and evaluate our segmentation nets. This dataset will be made public which may attract more research groups entering this field.

- We present an approach to generate compact binary iris templates (Deep-IrisCodes) from DeepIrisNet2.
- We present approaches to create large scale training data from public iris datasets to train DeepIrisNet2.
- We evaluate our approach for challenging scenarios such as iris recognition in visible spectrum and under a mobile phone environment. We summarize experiments and demonstrate that proposed framework outperforms all of the state-of-the-art works on the public datasets; UBIRIS v2.0, ND-0405, MICHE-1, CASIA Interval v4.

The rest of the paper is organized as follows. Related work is given in section 2. Section 3 explains proposed iris segmentation CNN nets. Section 4, discusses about datasets and Section 5 about spatial transformer (ST) module. DeepIrisNet2 network details are given in section 6, experiments in section 7 and conclusion in section 8.

## 2. Related Work

The first complete and automated iris recognition system was presented by Daugman [3] in 1993. He proposed generation of IrisCode based on quantized Gabor phase information extracted from normalized iris region. Following the pioneer work of Daugman and Wildes [24], a lot of researchers proposed a variety of feature extraction approaches for iris recognition, which can be categorized as: approaches based on- texture analysis [27], intensity variation analysis [28], phase-based [3], zero crossing [24] etc. Some researchers also explored feature extraction approaches and corresponding iris recognition system which does not rely on the transformation of the iris pattern to polar coordinates or on highly accurate segmentation [16,17,18,25] to make the iris recognition feasible with unconstrained image acquisition conditions. Basically, these approaches are based on finding extreme points in scale space and filtering them to get stable ones, followed by extraction of local features of the images around these stable points and generation of local descriptors from these local features for matching. However, it is found that if sufficient feature points are not extracted from iris or if noise points are not excluded properly, the recognition performance is degraded.

Some iris recognition methods further learn high-level features on top of low-level hand-crafted features [7,26,29] for better accuracy. We can roughly see these methods as a two-layer model though the parameters of local filters are hand-crafted. More deeper models (#layers > 3) based on hand-crafted filters are rarely reported in the literature. Because the filters (or parameters) of each layer are usually designed independently by hand and the dynamics between layers are hard to handle by human observations. Therefore, learning the parameters of all layers from data is the best way out.

Though, the CNNs were proposed long before, considerable progress has been made in last three years. Motivated by ground-breaking results in computer vision obtained by using deep convolutional neural networks, few research works [1,42,43] have also tried deep learning approaches for iris recognition tasks and a significant improvement in iris recognition accuracy is reported.

## 3. Proposed Iris Segmentation Methodology

### 3.1. Bounding Box Detection Network

The proposed iris bounding box detection CNN is motivated by you only look once (YOLO) architecture [5]. The original YOLO network contains 24 convolutional layers followed by 2 fully connected layers. However, for our problem, we experimented with some modifications in the network and found a smaller and efficient network shown in Table 3. Our network consists of 18 conv. layers (CONV1 to CONV18) and total 3 fully connected layers (FC1 to FC3). FC1 is added to further reduce the number of parameters by down sampling the input.

The approach of YOLO is quite simple and fast because unlike region proposal or sliding window based approaches, it uses features from full images at training and test time. YOLO directly regresses on the bounding box locations and class probabilities. YOLO divides each image into W-by-W regions and then from each region it directly regresses to find K object bounding boxes and a score for each of the N classes. It regresses on 5 numbers for each bounding box; these are center x, center y, width, height and confidence for the bounding box. All the bounding boxes in a region will have one set of class scores N. Thus, network's output is a vector of $W \times W \times (5K+N)$ numbers for each image. Different than original YOLO, our network is trained for 2 class problem (iris and pupil); N=2 with W=11, K=2. The final prediction is a $11 \times 11 \times 12 = 1452$ tensor. We first pre-trained a network on Imagenet dataset [34]. Using weights of pre-trained model, two separate models are trained for iris object detection in VS and NIR iris images. Training is performed using our ground-truth datasets; IrisDSGD-BBox_VS & IrisDSGD-BBox_NIR (Section 4.1). The size of input images is resized to 448-by-448 pixels. In each case 80% images are used for training, 5% for evaluation and 15% for Testing. To augment the training data, we perform horizontal flipping and image scaling with a factor of {0.8,0.9,1.1}. The training is performed roughly for 1,00,000 iterations, with a batch size of 32 and is stopped when accuracy stopped increasing. Initially, the learning rate was set to 0.001 and the momentum is taken as 0.9 and a decay of 0.0005. Throughout training the learning rate was reduced three times by a factor of 10. To avoid overfitting we use dropout at FC1 and FC2 layers. Sample bounding boxes detected are shown in Fig. 3.

### 3.2. Pixel-wise Segmentation Network

Once the images are processed by bounding box CNN, iris ROI is extracted using iris bounding box information and resized to a fixed size of 100-by-100 pixels. It should be noted that, the ROI inside bounding box will generally contain unwanted pixels like sclera, eyelid etc. Hence to obtain more accurate iris texture, we propose a semantic pixel-wise segmentation network. The network contains an encoder network and a corresponding decoder network followed by a pixel-wise classifier. The approach is motivated from SegNet architecture in Badrinarayanan et. al. [4]. We performed some key modifications in the original network of [4] to obtain a more optimal network for our task to segment an image in 3 classes; iris, pupil and background. Our network is shown in Table 4. The encoder network contains 4 convolution layers and similarly, decoder network contains 4 convolutional layers. Each convolution layer in encoder and decoder networks is followed by batch normalization [37] and a ReLU non-linearity. In the decoder network, max-pooling indices are used to perform upsampling of the feature maps which retains high frequency details in the segmented images. Decoder's final output is given to a soft-max classifier to produce class probabilities for each pixel. A sample segmented image is shown in Fig. 3. Training of our networks is performed from scratch using IrisDSGD-Iris_Mask_VS and IrisDSGD-Iris_Mask_NIR dataset (Section 4.1) for VS and NIR images respectively.

### 4. Data Sets

To train iris segmentation nets, we build groundthruth datasets. For DeepIrisNet2, we train different types of models for NIR and VS iris representation. We created large scale datasets for NIR and VS iris images from publicly available iris datasets. Under default settings, input images to DeepIrisNet2 are prepared by bounding box detection (Sec. 3.1) followed by pixel level semantic segmentation (Sec. 3.2) and resizing to 100x100 pixels.

### 4.1. IrisDSGD dataset for Segmentation Networks

Availability of a large scale data set is one of the key factor behind success of deep learning on various problems [34]. However, for iris detection or segmentation such a large scale dataset suitable for CNN training is not publicly available. To solve this issue, we build a large scale hand labeled dataset named as Iris Detection and Segmentation Groundtruth Database (IrisDSGD). The statistics of IrisDSGD dataset are shown in Table 1 & 2. The dataset contains 4 subsets; **BBox_VS** and **BBox_NIR**: Groundtruth Iris and Pupil Bounding boxes information for VS images and NIR images created by manually placing bounding boxes around pupil and iris region, **Iris_Mask_VS** and **Iris_Mask_NIR**: Groundtruth iris masks containing a manually created label for every pixel as iris/pupil or background.

Table1: IrisDSGD: Visible Spectrum Data Statistics

| Visible Spectrum Ground truths | Bounding Box (BBox_VS) | | Iris_Mask_VS | |
|---|---|---|---|---|
| | #Identities | #Irises | #Identities | #Irises |
| UBIRIS v2.0 | 200 | 6000 | 200 | 7000 |
| MICHE-1 | 70 | 3000 | 70 | 1000 |

Table2: IrisDSGD: NIR Data Statistics

| NIR Ground truths | Bounding Box (BBox_NIR) | | Iris_Mask_NIR | |
|---|---|---|---|---|
| | #Identities | #Irises | #Identities | #Irises |
| ND-0405 | 300 | 9000 | 300 | 9000 |
| CASIA v4 Interval | 150 | 1000 | 150 | 1000 |

Table 3: Architecture of Iris Bounding Box Detection Net

| Name | Type/Kernel/Stride/pad | Output |
|---|---|---|
| CONV1 | Convolution/7x7/2/1 | 224×224×64 |
| POOL1 | Max pooling/2×2/2 | 112×112×64 |
| CONV2 | Convolution/3x3/1/1 | 112×112×192 |
| POOL2 | Max pooling/2x2/2 | 56×56×192 |
| CONV3 | Convolution/1x1/1/1 | 56×56×128 |
| CONV4 | Convolution/3x3/1/1 | 56×56×256 |
| CONV5 | Convolution/1x1/1/1 | 56×56×256 |
| CONV6 | Convolution/3x3/1/1 | 56×56×512 |
| POOL3 | Max pooling/2x2/2 | 28×28×256 |
| CONV7 | Convolution/1x1/1/1 | 28×28×256 |
| CONV8 | Convolution/3x3/1/1 | 28×28×512 |
| CONV11 | Convolution/1x1/1/1 | 28×28×512 |
| CONV12 | Convolution/3x3/1/1 | 28×28×1024 |
| POOL4 | Max pooling/2x2/2 | 14×14×1024 |
| CONV13 | Convolution/1x1/1/1 | 14x14×512 |
| CONV14 | Convolution/3x3/1/1 | 14x14×1024 |
| CONV15 | Convolution/3x3/1/1 | 14x14×1024 |
| CONV16 | Convolution/3x3/2/1 | 7x7×1024 |
| CONV17 | Convolution/3x3/1/1 | 7x7×1024 |
| CONV18 | Convolution/3x3/1/1 | 7x7×1024 |
| FC1 | Full connection | 1024 |
| DROP1 | Dropout (0.5) | |
| FC2 | Full connection | 4096 |
| DROP2 | Dropout (0.5) | |
| FC3 | Full connection | #classes (2) |

Table 4: Architecture of Semantic Segmentation Net

| | Name | Type/Kernel/Stride/pad | Output |
|---|---|---|---|
| Encoder Net | CONV1 | Convolution/7x7/1/3 | 100x100x64 |
| | POOL1 | Max pooling/2x2/2 | 50x50x64 |
| | CONV2 | Convolution/7x7/1/3 | 50x50x96 |
| | POOL2 | Max pooling/2x2/2 | 25x25x96 |
| | CONV3 | Convolution/7x7/1/3 | 25x25x128 |
| | CONV4 | Convolution/7x7/1/3 | 25x25x128 |
| | POOL3 | Max pooling/2x2/2 | 13x13x128 |
| Decoder Net | UPSAMPLE3 | Upsample/scale=2 | 25x25x128 |
| | CONV4_Decode | Convolution/7x7/1/3 | 25x25x128 |
| | CONV3_Decode | Convolution/7x7/1/3 | 25x25x128 |
| | UPSAMPLE2 | Upsample/scale=2 | 50x50x96 |
| | CONV2_Decode | Convolution/7x7/1/3 | 50x50x96 |
| | UPSAMPLE1 | Upsample/scale=2 | 100x100x64 |
| | CONV1_Decode | Convolution/7x7/1/3 | 100x100x64 |
| | CONV_Classifier | Convolution/1x1/1 | #classes (3) |
| | COST | Softmax | |

## 4.2. NIR Dataset for DeepIrisNet2

The ND-CrossSensorIris-2013[21] is used to train our DeepIrisNet2 net for NIR images. The database contains 29986 images from LG4000 and 116564 images from LG2200 dataset. This dataset contains 27 sessions of data with 676 unique subjects. The identities from both the datasets are merged to obtain a larger dataset. The left and right eye images are given different class labels. In addition, 4 translated versions and 3 scaled versions from 15% of random samples are also added in training set. The identities which contain less than 25 images are removed. Finally, we built a NIR dataset containing roughly 300000 images in training set and 25,000 images in validation set from 1352 class labels.

## 4.3. Visible Spectrum Dataset for DeepIrisNet2

To create a larger training set for visible spectrum images, we combined images from UBIRIS.v1 [15], Visible UTIRIS V.1 [39], and a part of UBIRIS.v2 database [22]. The UBIRIS.v1 database contains total images from 241 subjects. The UBIRIS.v2 database contains total 11102 images from 261 subjects. The images are highly challenging and are captured during multiple sessions in unconstrained conditions; at-a-distance, on-the-move. We roughly selected 8500 images from UBIRIS.v2 from 211 individuals from both right and left eyes demonstrated in 422 classes in total. The UTIRIS dataset [39] contains 840 images captured in 2 sessions from 79 persons. In combined dataset, the left and right images are assigned different class labels. Finally, in combined dataset, we obtained roughly 11154 images from total 820 classes. Further, we augmented the training data by jittering the images by applying 4 translation and 3 scale transformations on all the images and thus total, 145000 images are obtained, from which around 130000 are used for training and remaining for va1idation.

## 5. Spatial Transformer (ST)

In [8], Jaderberg et al. introduced spatial transformer (ST) module that can be inserted into a standard neural network to perform the spatial manipulation of data within the network. Figure 1 shows spatial transformer module with its components. The architecture of an ST module has three main blocks: **i) Localization Network** - It takes the input feature as an input and through a number of hidden layers outputs the parameters ($\theta$) of the transformation ($\mathcal{T}_\theta$) to be applied to the input feature map. The size of $\theta$ depends upon the transformation type that is parameterized, such as for an affine transformation $\theta$ is 6-dimensional. **ii) Grid Generator**- it generates a grid ($G$) of coordinates in the input image corresponding to each pixel from the output image. **iii) Sampler Unit**- It takes the feature map as input and creates transformed output image sampled from the input at the grid points.

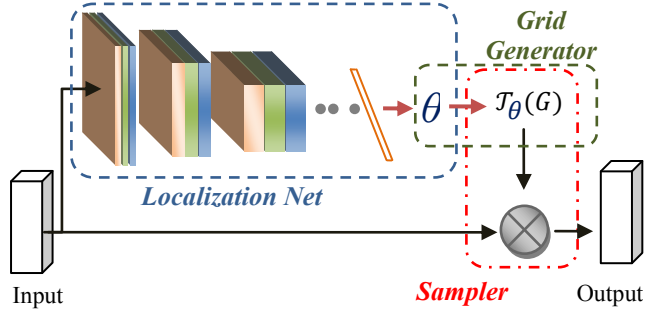

Figure1: The architecture of spatial transformer module

The ST module can perform an explicit geometric transformation of input images appropriately, including by selecting the most relevant region for the task. ST can learn to translate, crop, rotate, scale, or warp an image based on transformation parameters computed by a localization network. The parameters of the transformation are learnt in an end-to-end fashion using standard backpropagation algorithm without any need for extra data or supervision. Another advantage of ST is that, by attending to the most relevant region for the task, the ST also allows the further computation to be devoted to those regions. In [8], authors have shown learning meaning transformations using ST modules for fine-grained bird classification, and digit classification.

In our DeepIrisNet2, we utilize ST modules to predict the coefficients of an affine transformation e.g. translation, scale, rotation in the iris images. Thus the $\mathcal{T}_\theta$ is 2D affine transformation $A_\theta$. The regular grid in STN, $G = G_i|G_i = \{x_i^{out}, y_i^{out}\}$ provides coordinates in the input image corresponding to each pixel from the output image. If we take (x,y) as normalized coordinates such as, $(x, y) \in [-1,1] \times [-1,1]$, an affine transform ($A_\theta$) can be given as -

$$\begin{pmatrix} x^{in} \\ y^{in} \end{pmatrix} = \mathcal{T}_\theta(G_i) = A_\theta \begin{pmatrix} x^{out} \\ y^{out} \\ 1 \end{pmatrix} = \begin{bmatrix} \theta_{11} & \theta_{12} & \theta_{13} \\ \theta_{21} & \theta_{22} & \theta_{23} \end{bmatrix} \begin{pmatrix} x^{out} \\ y^{out} \\ 1 \end{pmatrix} \quad (1)$$

Pixel values in output feature map can be computed as-

$$I^{out}(x^{out}, y^{out}) = I^{in}(x^{in}, y^{in}) = I^{in}(\mathcal{T}_\theta(x^{out}, y^{out})) \quad (2)$$

The value of $\mathcal{T}(x^{out}, y^{out})$ is generally not an integer value on image grid, thus in our case we used bilinear interpolation to interpolate it.

## 6. DeepIrisNet2 Architecture and Training

The proposed DeepIrisNet2 network is carefully designed, taking in consideration the suggestions in recent literature and observing other highly successful architectures. We tried various configurations and our final network configuration is shown in Fig. 2. The architecture of DeepIrisNet2 is quite different from the other network architectures proposed in literature for various tasks. The DeepIrisNet2 configuration not only achieves improved accuracy but it also converges faster. We trained separate networks for NIR and VS iris recognition using NIR and visible spectrum datasets as given in section 4. The networks are named as DeepIrisNet2-NIR and DeepIrisNet2-VS for NIR and visible spectrum images respectively. The two networks are same except the first layer. The input to our networks is fixed-size 100-by-100 pixel images. During training, the NIR and VS images are preprocessed by subtracting the mean Gray and RGB value respectively, computed on the training sets, from each pixel. The network contains 10 convolution layers with 5-by-5 kernel size for first and 3-by-3 kernel sizes for rest of the conv. layers. We also perform stacking of conv. layers at certain intermediate layers in the network without spatial pooling in between such as conv3-conv4, conv5-conv6, conv7-conv8-conv9. By removing down sampling operation of input at these layers, the network is able to learn more complex higher dimensional representations. It is very common practice to take two to three fully connected (FC) layers with dropout at classifier level. This design in classifier level is adopted by most of the networks proposed in literature for various tasks. In our DeepIrisNet2, empirically we adopted a different approach and rather than using multiple FC layers in classifier level, we added two conv. (i.e. Conv9 and conv10) with dropout regularization (dropout ratio set to 0.5) on each of them. Dropout helps in minimizing overfitting. We observed that this design is less prone to overfitting. The output of last conv. layer i.e. conv10 is fed to a C-way Softmax (where C is the number of classes) which produces a distribution over the class labels. Throughout the whole network, the spatial pooling is carried out by Max-pooling, except for Pool5, which is average pooling. We use ReLU activation function after all convolution layers, except for conv10. The weights in convolutional and linear layers are initialized from scratch using random Gaussian distributions. We trained our network with a mini batch size of 256, and momentum set to 0.9. The training was regularized by weight decay (the L2 penalty multiplier set to $5 \times 10^{-4}$). The learning rate was initially set to 0.01, and then decreased by a factor of 10 when the validation set accuracy stopped improving. There was no cap on the number of epochs and training was stopped when the validation error stopped improving for a good amount of epochs. In total, the learning rate was decreased 3 times. Since the network of our DeepIrisNet2 is quite deep, we experimented by adding auxiliary classifiers at intermediate layers as used in some previous works [32]. Empirically, we found that in case of our network, best result are obtained by connecting auxiliary classifiers to Pool1, Pool2, Pool3, and Pool4 layers. This way total four auxiliary classifiers which are shallow convolutional neural networks are incorporated in our network. The configurations of different auxiliary classifiers are different and details are shown in Fig. 2. We observed various benefits of adding auxiliary classifiers such as, a) increase in gradient signal that is propagated back in the network to combat the vanishing gradient problem with very deep architectures, b) more discriminative features at lower stages (we tested with features extracted from pool5 stage), c) additional regularization in the network which makes the network more difficult to overfit. During training of network, the loss obtained from these auxiliary classifiers is added into the network loss from main branch with weight 0.3.

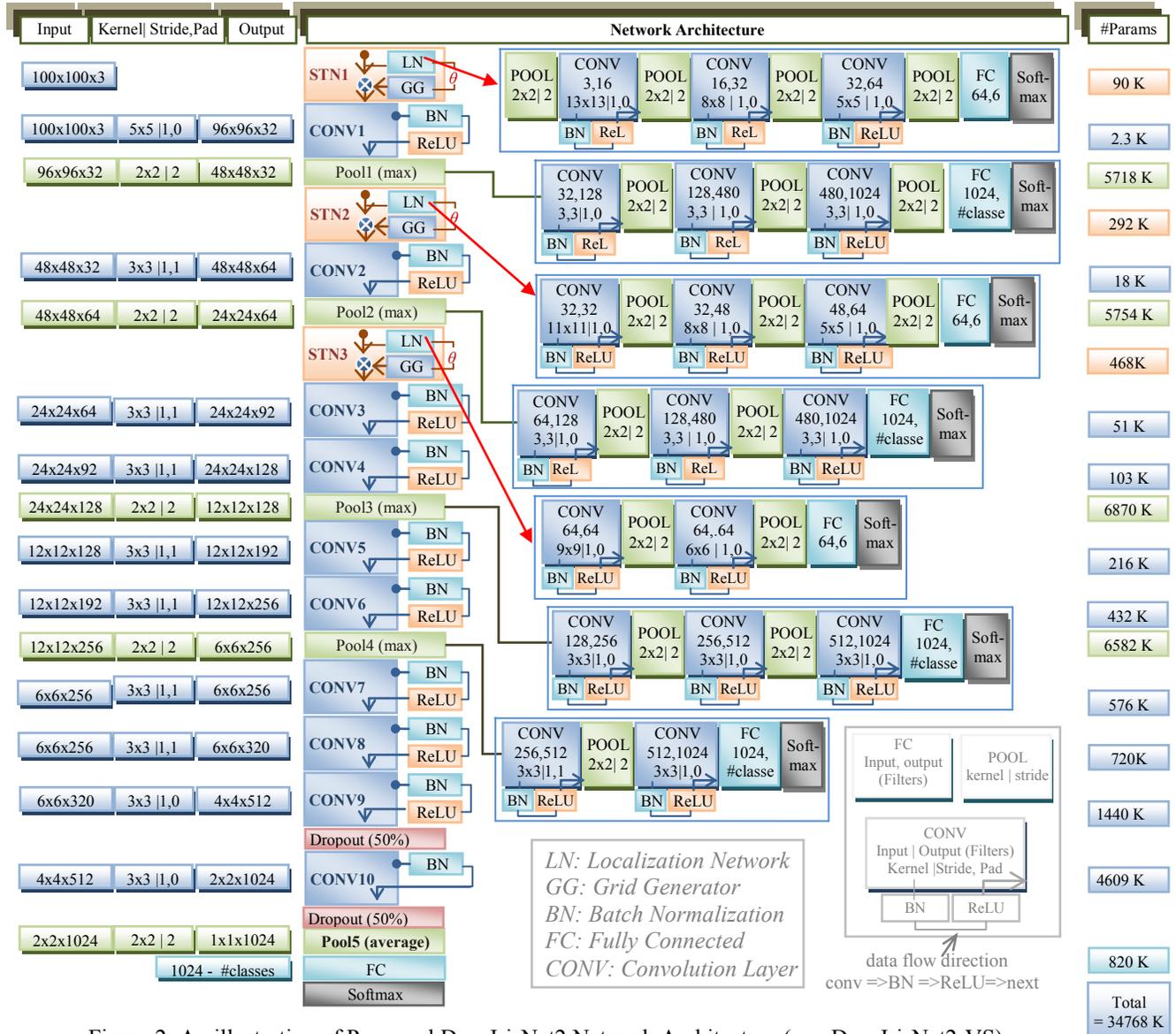

Figure 2: An illustration of Proposed DeepIrisNet2 Network Architecture (e.g. DeepIrisNet2-VS)

The pooling operations in CNNs are performed mainly to obtain sub-sampling of the input feature map. The pooling is done over local neighborhoods with no overlap and it creates summaries of each sub-region. Thus it also provides little bit of positional and translational invariance. However, due to pre-defined pooling mechanism and small spatial support for max-pooling (e.g. 2-by-2 pixels) this spatial invariance is not very effective to large variations in input feature maps.

To obtain spatial invariance against large variations and in dynamic manner, in our DeepIrisNet2 architecture, we incorporated multiple spatial transformers modules [8], which are capable of generating automatic transformation of input feature maps as given in section 5 above. In particular, in our network, the first spatial transformer module is inserted immediately after the input layer and two more spatial transformers modules are added after Pool1 and Pool2 layers. The networks of different spatial transformers modules are different and full details of various spatial transformers modules embedded in DeepIrisNet2 is given in Fig. 2. When number of channels in input to spatial transformers module are more than one, the same transformation is applied on all the channels. By using multiple spatial transformers in DeepIrisNet2, we are able to obtain transformations of increasingly abstract representations.

Table 5: Accuracy of Bounding Box Detector CNN

| Dataset | #Input Images | #Correct bounding box detections | | | |
|---|---|---|---|---|---|
| | | #Irises | Accu.(%) | #Pupil | Accu.(%) |
| ND-0405 [21] | 50000 | 49746 | 99.49 | 49762 | 99.52 |
| MICHE [23] | 1000 | 966 | 96.60 | 991 | 99.10 |
| UBIRIS.v2 [22] | 3000 | 2941 | 97.80 | 2970 | 99.01 |
| LG2200 [21] | 116564 | 115643 | 99.21 | 116563 | 100 |
| LG4000 [21] | 29986 | 29895 | 99.69 | 29985 | 100 |
| Casia v4 Interval [20] | 1000 | 993 | 99.40 | 999 | 99.95 |
| UBIRIS V.1 [15] | 1877 | 1870 | 99.69 | 1876 | 99.99 |
| UTIRIS [39] | 824 | 822 | 99.80 | 824 | 100 |

Table 6: Accuracy of Semantic Segmentation CNN

| Dataset | R | | P | | F-measure | |
|---|---|---|---|---|---|---|
| | μ | σ | μ | σ | μ | σ |
| IrisDSGD-Iris_Mask_VS | 96.45 | 1.84 | 96.89 | 1.93 | 96.98 | 1.88 |
| IrisDSGD-Iris_Mask_NIR | 98.01 | 1.91 | 97.56 | 1.79 | 97.78 | 1.84 |

The DeepIrisNet2 contains 11 layers with parameters without spatial transformers modules. With spatial transformers modules it contain 14 layers and if we take pooling layers also, it contains 20 layers. If we count each and every building blocks used in construction of DeepIrisNet2, then total numbers of layers are around 100. The total number of parameters in the network are ~34M. The proposed iris recognition pipeline based on our approaches is shown in Fig. 3.

## 7. Experiments Analysis

### 7.1. Performance Analysis of Segmentation Nets

In case of bounding box detection CNN, we used 15% of the ground truth images for testing purpose which are not used during training. First, we computed accuracy by using area under a receiver operating characteristic (ROC) curve, also known as *AUC*. Bor BBox_VS test set (~1350 images), we obtain an accuracy of 96.78% and for BBox_NIR test set (~1500 images), the accuracy reported is 97.87%, which are quite impressive. Second, we used this CNN to detect iris object on various datasets (without ground truth) and computed detection accuracy using visual inspection. The accuracies are reported in Table 5.

In case of semantic pixel-wise segmentation network, the CNN's iris mask are compared with ground truth mask and we report here, the mean (μ) and standard deviation (σ) for precision (P), recall (R) and F-measure (F) metrics [9] obtained over complete held-out test set. The results are highly promising and are shown in Table 6.

Since, our iris segmentation pipeline is completely different than all previous methods, we are unable to show a comparative analysis with other existing method.

Table 7: EER (%) Analysis on MICHE database

| Approach | IP5_I_F | IP5_I_R | GS4_I_F | GS4_I_R | IP5_O_F | IP5_O_R | GS4_O_F | GS4_O_R |
|---|---|---|---|---|---|---|---|---|
| Daugman [3]* | 3.74 | 8.35 | 8.399 | 6.16 | 6.78 | 10.41 | 12.45 | 10.52 |
| Masek [10]* | 17.01 | 13.90 | 18.83 | 17.26 | 21.86 | 24.01 | 21.22 | 20.83 |
| Ma et al. [28]* | 17.01 | 12.84 | 17.10 | 18.68 | 22.01 | 29.98 | 20.60 | 20.63 |
| Ko et al. [40]* | 14.58 | 11.29 | 14.06 | 14.68 | 18.11 | 21.78 | 17.70 | 18.05 |
| Rathgeb and Uhl [36]* | 21.05 | 20.83 | 21.13 | 25.68 | 27.43 | 28.12 | 27.30 | 24.39 |
| Rathgeb and Uhl [35]* | 26.84 | 26.31 | 30.43 | 33.17 | 26.76 | 30.39 | 31.09 | 26.04 |
| Raghavendra et al. [41]* | 2.07 | 0.48 | 4.23 | 4.18 | 7.82 | 8.57 | 8.11 | 6.29 |
| Raja et al. [26]* | 0.02 | 6.25 | 2.50 | 3.96 | 4.18 | 6.25 | 6.27 | 2.06 |
| MIRLIN ** | 2.73 | 2.84 | 3.33 | 3.16 | 3.33 | 1.12 | 5.98 | 3.32 |
| **Proposed** (DeepIrisNet2-VS) | 1.98 | 1.56 | 2.48 | 2.33 | 2.65 | 1.05 | 3.98 | 1.67 |

*Score is obtained from [26]    **Score is obtained from [33]*

## 7.2. Performance Analysis of DeepIrisNet2

We train two types of models named DeepIrisNet2-VS and DeepIrisNet2-NIR for visible spectrum and NIR iris images, respectively. For iris representation purpose, we use features extracted from Pool5 layer with dimension 1024. The similarity score is computed using cosine distance. To report unbiased comparisons, some results on the test datasets used in this work are taken directly from the published work. Under default settings, all of the images in test datasets are processed by the same pipeline as the training dataset and normalized to 100-by-100 pixels. The jittering is not applied on test images.

**Experiment 1: Iris Recognition for VS iris images**

To empirically evaluate the capabilities of proposed DeepIrisNet2 under visible spectrum iris capturing, we perform performance analysis testing on challenging UBIRIS.v2 and MICHE datasets. The test set for UBIRIS.v2 contains ~2300 images with 100 class labels, which are kept left out of training/validation set. The MICHE-I dataset contains images captured using iPhone5, SamsungGalaxyS4 and SamsungGalaxyTab2. The images are captured using both frontal and rear camera in indoor and outdoor conditions. The major purpose of the MICHE database is to assess, in general, the performance of iris recognition algorithms on mobile devices using their built-in camera. During evaluation of MICHE, we partially followed the experiments speculated by Raja et al. [26]. Also, following [16,44], Samsung GalaxyTab2 images are not considered for evaluation. In MICHE test set around 1200 images from iPhone5 and 1200 images of SamsungGalaxyS4 set are taken and these images are further divided into four subcategories: front/rear and outdoor/ indoor. Finally, 50% (~150) images are taken into gallery set and remaining 50% (~150) images into probe set for each category. The results (EER in %) are given in Table 7. Since our network is trained on different database, it is easily observed that the proposed DeepIrisNet2-VS network not only generalizes well to new database but also provides good recognition rate. It should also be noted that, we used almost all the images of MICHE in each category for testing (total ~2500 from 75 subjects), whereas previous works [16,44] utilized only

Table 8: EER (%) Analysis on UBIRIS.v2 database

| Approach | # irises | #images | EER% |
|---|---|---|---|
| SIFT [31]* | 181 | 1000 | 25.21 |
| Daugman [31]* | 181 | 1000 | 31.06 |
| SRLRT[30]* | 152 | 904 | 24.56 |
| Monogenic Log-Gabor [30]* | 152 | 904 | 27.39 |
| Log Gabor [30]* | 152 | 904 | 36.45 |
| QSOC[19]* | 152 | 904 | 23.10 |
| LECM [15]* | - | 1000 | 18.09 |
| WCPH[14]* | 78 | 503 | 22.78 |
| Log Gabor [13]* | 152 | 904 | 27.45 |
| 2D Gabor filters [12]* | 114 | 1227 | 26.14 |
| Global+local ZMs phase [11]* | 151 | 864 | 11.96 |
| DeepIrisNet2-VS** | 100 | 2300 | **8.51** |

Table 9: EER (%) Analysis of "DeepIrisNet2-NIR"

| Approach | ND-0405 | | CASIA Interval V4 | |
|---|---|---|---|---|
| | #Images | EER | #Images | EER |
| Ko [40] | 2000 | 12.90 | 2639 | 1.20 |
| Ma [28] | 2000 | 9.12 | 2639 | 0.98 |
| Masek [10] | 2000 | 14.72 | 2639 | 8.75 |
| DeepIrisNet2-NIR** | 2000 | 1.48 | 2639 | 0.29 |
| DeepIrisNet2-NIR-BBOX ** | 2000 | 1.48 | 2639 | 0.30 |
| DeepIrisNet2-NIR-normalized** | 2000 | 1.47 | 2639 | 0.28 |
| DeepIrisNet2-NIR (Binary) DeepIrisCode** | 2000 | 2.11 | 2639 | 1.25 |

*Score reported in Publication    **Proposed Approach*

limited (~50-100) images to report accuracies.

The performance analysis (EER) on UBIRIS.v2 database is presented in Table 8. Table 8 also illustrates the performance comparison between various algorithms. It is clearly seen that, the proposed DeepIrisNet2-VS is quite effective. We also observe that the proposed network outperforms all previous methods with a huge margin.

**Experiment 2: Iris Recognition for NIR iris images**

The testing is performed on well known challenging NIR datasets; ND-iris-0405, Casia Interval V4 [20]. The testing datasets are completely different than the datasets used for network training. The ND-iris-0405 contains 64980 images from 356 subjects; Casia Interval V4 contains total 2639 images from 249 subjects. For comparative analysis, we also evaluated well known Ko et al. [28][1], Ma et al [38][1], Masek [10] approaches. The EER of the approaches is reported in Table 9. In summary, the proposed DeepIrisNet2-NIR architecture achieves significant improvement over strong baseline for iris recognition algorithms on ND-iris-0405 and CASIA Interval V4 databases, which further confirm the effectiveness of proposed DeepIrisNet2 for NIR images.

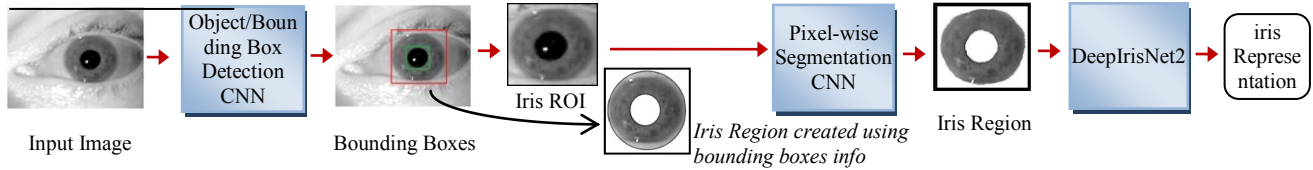

Figure 3: Illustration of Proposed Iris Recognition Pipeline

**Experiment3: DeepIrisNet2 with Poorly Segmented Iris Images (Segmentation with Bounding Box data)**

This experiment is setup to study the effect on iris recognition accuracy of perfect iris segmentation vs. poorly segmented iris region and ST modules. For this purpose, a separate network is trained from scratch using poorly segmented NIR dataset of section 4.2. For this, in input images, pupil and other unwanted (noise) regions are masked using information provided by Bounding Box CNN (Sec. 3.1). In the process, the pupil is masked using circular region computed by taking center of pupil bounding box as center and average of height and width as diameter. Similarly, iris outer region is masked using iris bounding box information. Sample masked image is shown in Fig. 3. We name this model as DeepIrisNet2-NIR-BBox. During training of this network, we observed that the network converges faster when training is performed on perfectly segmented images, but after around 1000 iterations, the training and validation errors of both the models were almost same. The performance (EER) of DeepIrisNet2-NIR-BBox is reported in Table 9, and it is illustrated that the verification rate reported by DeepIrisNet2-NIR and DeepIrisNet2-NIR-BBox are almost same. The surprising performance of DeepIrisNet2-NIR over poorly segmented images can be attributed to introduction of Spatial Transformer (ST) modules in the network. To verify it, we trained two DeepIrisNet2-NIR networks without ST modules. In case of perfectly segmented irises there was a minor reduction (0.35%) in the accuracy, however, in case of poorly segmented irises difference in accuracy was around 1%.

**Experiment 4: Non-Normalized Vs. Normalized images**

This experiment is designed to perform comparative analysis between the DeepIrisNet2-NIR network trained on iris images segmented by our segmentation pipeline (Section 3) and DeepIrisNet2-NIR network trained on iris images normalized using Daugman's Rubber Sheet model. To perform normalization, we localized iris boundaries used IrisSeg framework [9] and during normalization, the images are unwrapped and transferred in a rectangular image in polar coordinates of size 50-by-200 pixels. Normalized images are further mapped to 100-by-100 pixels images following the approach adopted in [1] (Fig. 4). The two DeepIrisNet2-NIR models trained on normal images and normalized images are named as DeepIrisNet2-NIR and DeepIrisNet2-NIR-normalized. The accuracy of this network is reported in Table-9. It is found that, the network trained using normalized images converges faster and performs little better.

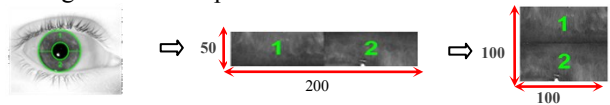

Figure 4: Normalized iris mapping to 100x100 pixels

**Experiment 5: Real-valued Features Vs. Binary Feature Vector (DeepIrisCode)**

To generate binary iris codes, we consider the feature activations at the last layer, Pool5 layer with 1024 dimensional tensor output. We then convert these activations into binary codes by thresholding by a value zero. The thresholding produces sparse vector, therefore such representation generally degrades the performance. It should be noted that, to obtain dense and compact real valued output at Pool5, we have not applied ReLU at conv10. As illustrated in Table 9, the binary representation of pool5 features from DeepIrisNet2-NIR, sacrifices ~0.5 to 1% accuracy by directly calculating the Hamming distances. However, Binary code is economic for storage and fast for iris matching.

## 8. Conclusion

In this work, we have made multiple contributions; i) we presented a Iris/Pupil bounding boxes detection CNN and another CNN architecture for semantic segmentation of iris texture, ii) we build a large scale dataset named IrisDSGD containing hand labeled annotations for iris, pupil bounding boxes and pixel level labeling of iris mask, iii) we presented a robust deep CNN architecture named DeepIrisNet2, which is shown to work effectively without Daugman's iris normalization step and without perfect iris segmentation. Through extensive experimental analysis, it is shown that, proposed approach generalize well to new images as well as to new subjects. On very large and highly challenging datasets, it is shown to perform competitively, obtains breakthrough accuracy and achieves new state-of-the-art recognition rate. Future work will be focused to train an end-to-end iris recognition solution.